# Doctor's Handwritten Prescription Recognition System In Multi-Language Using Deep Learning

Pavithiran G[1], Sharan Padmanabhan[2], Nuvvuru Divya[3], Aswathy V[4], Irene Jerusha P[5], Chandar B[6]
Department of Computer Science and Engineering,
Panimalar Institute of Technology, Chennai, India
pavithiran.nasa@gmail.com[1], sharanmozart@gmail.com[2], divyanuvvuru640@gmail.com[3],
aswathy3124@gmail.com[4], irene.jp2002@gmail.com[5], chandar.barathi123@gmail.com[6]

*Abstract*—Doctors typically write in incomprehensible handwriting, making it difficult for both the general public and some pharmacists to understand the medications they have prescribed. It is not ideal for them to write the prescription quietly and methodically because they will be dealing with dozens of patients every day and will be swamped with work.As a result, their handwriting is illegible. This may result in reports or prescriptions consisting of short forms and cursive writing that a typical person or pharmacist won't be able to read properly, which will cause prescribed medications to be misspelled. However, some individuals are accustomed to writing prescriptions in regional languages because we all live in an area with a diversity of regional languages. It makes analyzing the content much more challenging. So, in this project, we'll use a recognition system to build a tool that can translate the handwriting of physicians in any language. This system will be made into an application which is fully autonomous in functioning. As the user uploads the prescription image the program will pre-process the image by performing image pre-processing, and word segmentations initially before processing the image for training. And it will be done for every language we require the model to detect. And as of the deduction model will be made using deep learning techniques including CNN, RNN, and LSTM, which are utilized to train the model. To match words from various languages that will be written in the system, Unicode will be used. Furthermore, fuzzy search and market basket analysis are employed to offer an end result that will be optimized from the pharmaceutical database and displayed to the user as a structured output.

*Keywords*—*Convolutional Neural Network(CNN), Recurrent Neural Network(RNN), Long short-term memory (LSTM), Fuzzy Search, market basket analysis, Optical character recognition (OCR), Unicode, application, Image pre-processing*

## I. INTRODUCTION

The art of handwriting allows each person to convey their ideas on paper in their own unique way. Depending on the individual,it might vary greatly. Specifically, when talking about a doctor's busy schedule, more consultations are scheduled in a short amount of time, and the diagnosis is given more importance than the prescription's handwriting. As a result, they frequently have poor handwriting, making it sometimes difficult to read the prescription and recognise the drugs and their possible dosages.It is quite challenging for patients and young pharmacists to distinguish the doctor's handwriting.The spectrum of outcomes from drug errors ranges from no apparent symptoms to death. It can sometimes result in a new ailment that is either transient or permanent, such itchiness, rashes, or skin deformity. Despite being rare, drug mistakes can seriously harm individuals. According to an Egyptian study[2], almost 96 percent of the public supported a software application that could be used to translate a doctor's handwriting into digital text. In most cases, doctors only indicate the type of medicine, such as tablet, capsule, or syrup, using acronyms and short forms. There are systems that have been suggested that can employ the Deep Convolutional RNN approach to recognise the alphabet and numerals from a written text in English[1]. India is widely renowned for having a wide variety of cultures and languages. Depending on the patient's needs, doctors may occasionally refer to the prescription drugs in regional languages. So, the application will be far more widely utilized by both pharmacists and regular people with regional language support.

The primary purpose of this study is to create an application that can actively recognise medical prescription images or scan them for subsequent conversion to digital text. This is accomplished by deploying deep learning techniques such as CNN, RNN, and LSTM for image recognition. In order to process the data OCR, word segmentation is used in image processing. Furthermore, a tailored output is provided to offer an optimal summary that most consumers may grasp even if they have no prior knowledge. The construction procedure is far more convenient for average people to use in order to take their daily dosages as prescribed by their doctors. This also makes it easier for new pharmacists and also consumers to conduct their tasks more efficiently and properly.

## II. REVIEW OF RELATED LITERATURE AND STUDIES

1. *Recognition of Doctors' Cursive Handwritten Medical Words by using Bidirectional LSTM and SRP Data Augmentation.(2021)*

The paper suggests an online handwritten recognition system to identify doctors' handwriting and create a digital prescription using machine learning techniques. The study developed a primary "Handwritten Medical Term Corpus" dataset with 17,431 data samples comprising 480 words from 39 Bangladeshi doctors. On the preprocessed pictures, a new data augmentation technique called SRP is used to increase the number of data samples. Following this, a sequence of line data is extracted from both the original and augmented image data. Bidirectional LSTM is applied to the sequential line data derived from the augmented handwritten images to produce complete end-to-end recognition. The model achieved 73.4% accuracy without data expansion and 89.5% accuracy with SRP data expansion.

2. *Handwriting Recognition for Medical Prescriptions using a CNN-Bi-LSTM Model(2021)*

It is difficult to decipher a doctor's handwriting on a prescription. In this paper, they used neural network techniques such as CNN and BI-LSTM for predicting doctor's handwriting from medical prescriptions. The CTC loss function is used for normalization. This model builds on the IAM dataset. Image acquisition and data augmentation are used for image preprocessing. Furthermore, it is passed as input to 7 convolution layers of a neural network. 32 training epochs were used by the training model, which took six hours to complete training and, on a graph, loss values are represented.

3. *Medical Handwritten Prescription Recognition Using CRNN.(2019)*

The approach established a Convolutional Recurrent Neural Network (CRNN) technology using Python that can interpret handwritten English prescriptions and translate them into digital text. For this, datasets with 66 different classes, including alphanumeric characters, punctuation, and spaces, were used. Since prescriptions generally contain two or three words, the training was carried out using short texts. Normal handwriting and prescriptions from doctors were used to train the model. The system got a 98% accuracy rate after taking training time and data input into account. This paper further stated that in order to enhance the results, more work is needed on input handling techniques.

4. *Intelligent Tool For Malayalam Cursive Handwritten Character Recognition Using Artificial Neural Network And Hidden Markov Model. (2017)*

The approach uses the Hidden Markov Model (HMM) to recognise cursive handwritten Malayalam characters. By employing a median filter, the algorithm used here helps to avoid errors caused by noise in the scanned image. Furthermore, Artificial Neural Network (ANN) aids in the acquisition of better classification and provides the best matching class for input. The samples used are of high quality in order to reduce the complexity of the recognition process. This method yields better results in terms of speed and accuracy. As a result, the combination of both English and Malayalam characters can be recognised as a future work.

## III. DATA COLLECTION

Data is a crucial component of this system for training and testing the deep learning model. As a result, for languages where there is no existing data, the data will be created from scratch. This will be accomplished by physically acquiring the data by scanning physical materials. By doing so, we can retain maximum accuracy in the recognition model, increasing its dependability. However, there is another method by which we may generate a complete prescription data set from scratch by translating the English prescription using Google API and then producing handwritten text in other languages using GAN. However, there will be a loss of precision using this procedure[3].

## IV. DESIGN AND METHODOLOGY

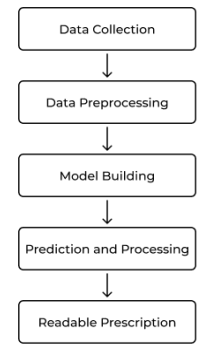

Fig.1 Phases of design

As shown in Fig.1, these are the methods used to process a physician's handwritten note. It's been discussed shortly below:

A. Data pre-processing and training model:

- Data Preprocessing:
  Following data collection, the gathered data will be unsorted so processed before training the model. We all know that if we feed the model garbage data, we will get garbage results. As a result, we must tread cautiously when training the model. So, the following data pre-processing sets will be taken:

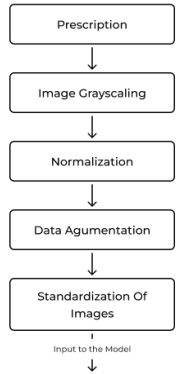

Fig.2 Data pre-processing

- Grayscale conversion:

To prepare the pictures for model training and drawing a conclusion from them, image preprocessing is performed. Gray scaling is the first step in digital image pre-processing that must be done. Here each pixel's value solely encodes the light's intensity information.

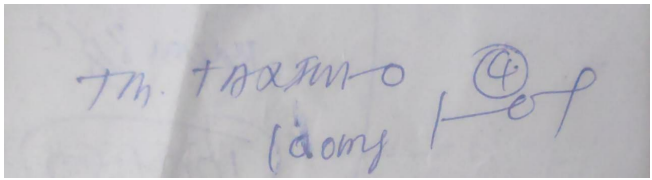

Fig:3 Original Image

There are just three colors used in grayscale images: black, white, and gray, which comes in a variety of tints. The digitalized images of the prescriptions are gray scaled first. The photos in grayscale are then prepared for the word segmentation procedure.

In Figure 3, the original digital image is converted into a grayscale image which possesses only black, white and shades of gray color as mentioned above and a later word segmentation process is carried out[4].

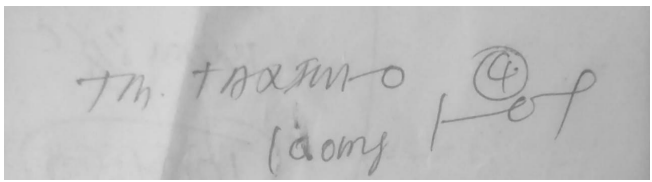

Fig:4 .Gray Scaled Image

Utilizing blank spaces and other special characters, the prescription's words are divided into segments. After segmentation, the words are normalised. The gray scaled pictures' initial pixel values range from 0 to 255. We divide the value of the gray scaled pixel by 255 in order to conduct normalization, which is the conversion of the pixel range from 0 to 1.

After normalization, we go for the edge detection step which is a fundamental step for image preprocessing. Here, we make an attempt to identify the significant properties such as discontinuities in the photometrical, geometrical, and physical attributes of the objects in the image and try to localize these variations. The edge detector accepts discrete, digitized images as input and produces an edge map as output. We apply smoothing to reduce and suppress image noise.

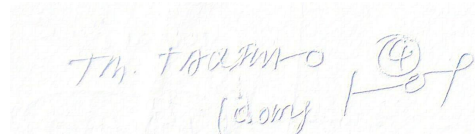

Fig: 5 Edge Detected Image

The figure 5 depicts the original image after the edge detection process is accomplished. This image is then set for a smoothing process so that the unnecessary image noise would be suppressed.

- Normalization:
It is done to resize the picture pixels to a preset range where they will be consistent with the data given. And by doing so, the model will perform common learning, avoiding irregularities in training the model.

- Data Augmentation:
It is used to make slight changes to a picture in order to deliver a diverse range of data in a single identifiable format. And the procedure frequently includes rotation, cropping, shearing, horizontal and vertical flipping, and so on. Furthermore, by doing so, we can keep the neural network from learning from irrelevant data. And followed by this image stabilization will be performed.

- Image standardization:
To convert the height and weight of a picture to a common scale, which will convert all of the given data into an appropriate size. We increase the consistency of the given data as well as the quality, which is a priority while completing these activities.

Once these processes are in place, there will be an optimal dataset that can be utilised to train the model. As a result, the neural network model will not be trained on irregular data[5].

- Training the model:

After processing the handwritten text input, we must train the model using structured data. So, in order to get the optimal recognition model, we need to avoid overtraining the model, which will result in the model being suited only to that particular data. To circumvent this, we will train the model using 50 ecop, which is the appropriate ratio for training the model to recognise the presented information.

B. Model Building and Prediction:

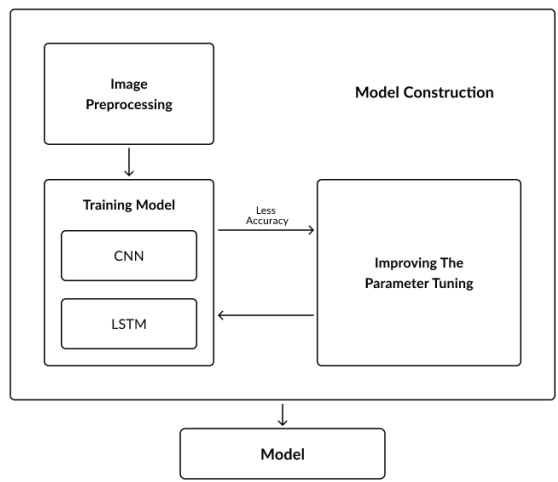

Fig.6 Model working

Neural networks are used in every area of the system. CNN, RNN, and LSTM neural networks are used to train the model. The Python libraries TensorFlow and Keras are used for coding. As previously stated, the model has two CNN layers and two RNN layers. 32 and 64 filters are used for the two levels of CNN, respectively. RNN has two layers, each with 64 and 128 filters. The model is trained using these approaches. The SoftMax activation function is used in this technique. Connectionist Temporal Classification (CTC) Loss is used to compute the loss function. As we could see in Figure 6.

Furthermore, to avoid errors, check the medical database if the model recognises the few phrases of it or compare the medicine. Following the recognition process, we will additionally utilize market basket analysis and fuzzy search to offer further assistance to the model. In such case we can see the figure 7 for reference

- Fuzzy search:

It is used to deliver the precise predicted tablet word even if the model recognised only a few words in it, whereas fuzzy search would provide the name even if the spelling is incorrect. This is done by using the provided medicine database[6].

- Market basket analysis:

This is used to save time when we detect commonly used drug names and data; this may be used to fetch the name faster based on use. So, wew conserve the time for analyzing the words.

C. Unicode Data:

Everyone knows that OCR is the finest method for character recognition. Because Indian scripts are difficult to recognise, we are adding a post-processing step to the OCR to achieve valid character recognition. In this method, typical characters that have already been divided into meaningful parts and used to train the model with unicodes are combined to form a string using mapper. The result will then be classified as valid UAM (unicode Approximation Model) or invalid UAM. The invalid UAM is processed again with a string pattern matching algorithm and cross-checked against legitimate UAM databases, and the mapper alters the character or string accordingly to generate the correct UAM.

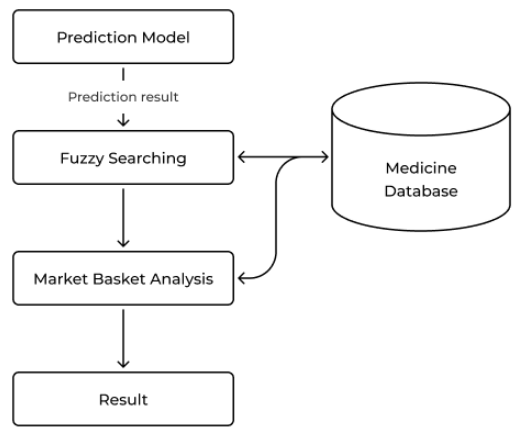

Fig. 7 Prediction Optimiser

D. Application interface:

The processed text will be given in an organized manner in the application interface, along with a brief explanation of the identified medicines. And this application will be autonomous and simple to use, allowing any user to effortlessly upload an image and receive understandable text in return.

V. RESULTS AND TEST CASES

To achieve the best possible outcome, we ensured that the model's data was correct. To get a favorable result, we trained the model with 50 ecop so that it is not accustomed to the training data and can predict the words.Furthermore, we divided the data into 90% for training and 10% for testing and validating the model. And we made certain that the CTC loss function was kept as low as possible in order to get the best possible word prediction. Finally, the predicted text from the image is revised using a medical data set using the aforementioned algorithms to offer a concise summary of the prescribed pill in the display.

However, in order to understand the real-world issues, we attempted to test our model at a local pharmacy shop to see whether it could distinguish diverse handwritten notes from physicians. Surprisingly, we are able to collect a slew of faults and benefits when running the test case. In a nutshell, we can detect a doctor's handwritten notes or prescriptions and deliver them to consumers in the language of their choice.

## VI. CONCLUSIONS AND FUTURE WORK

Finally, this application interface will make it simple for people to access the model and interact with it through the application. Furthermore, this technique allows the majority of users to verify the notes or prescriptions without any prior knowledge in calligraphy analysis. Therefore, this technology will eliminate human mistakes and pave the way for customers to assess it without the assistance of experts.

We might improve the accuracy more in the future by supplying more data for training. Furthermore, the method may be tweaked to produce results even faster. Make this programme cross-platform and long-lasting for even the most stringent requirements.